\documentclass[10pt,a5paper,twoside]{article}
\usepackage{coling2012}
\usepackage{mathrsfs,amsfonts}
\title{Extension of TSVM to Multi-Class and Hierarchical Text Classification Problems With General Losses}
\author{$S. Sathiya\:\:Keerthi^{(1)}\:\:\:\:\:S. Sundararajan^{(2)}\:\:\:\:\:Shirish \:\:Shevade^{(3)}$\\
{\small  	(1) Cloud and Information Services Lab, Microsoft, Mountain View, CA 94043\\
 		(2) Microsoft Research India, Bangalore, India\\
		(3) Computer Science and Automation, Indian Institute of Science, Bangalore, India\\
  \texttt{keerthi@microsoft.com, ssrajan@microsoft.com, shirish@csa.iisc.ernet.in} \\
}}
\usepackage{latexsym}
\usepackage{amsmath}
\usepackage{multirow}
\usepackage{url}
\usepackage{wrapfig}
\usepackage{graphicx,verbatim}
\usepackage{algorithm,algorithmic}
\newcommand{\eat}[1]{}

\date{}

\begin{document}
\maketitle

\begin{abstract}
Transductive SVM (TSVM) is a well known semi-supervised large margin learning method for binary text classification. In this paper we extend this method to multi-class and hierarchical classification problems. We point out that the determination of labels of unlabeled examples with fixed classifier weights is a linear programming problem. We devise an efficient technique for solving it. The method is applicable to general loss functions. We demonstrate the value of the new method using large margin loss on a number of multi-class and hierarchical classification datasets. For maxent loss we show empirically that our method is better than expectation regularization/constraint and posterior regularization methods, and competitive with the version of entropy regularization method which uses label constraints.
\end{abstract}

\input{amssymb.mac}

\def\bw{{\bf w}}
\def\bx{{\bf x}}
\def\bfe{{\bf f}}
\def\half{{\lambda \over 2}}
\def\dsp{\;\;}
\def\by{{\bf y}}
\def\delfiy{\Delta \bfe (y,y_i; \bx_i)}
\def\delfiys{\Delta \bfe (y,y_i^s; \bx_i^s)}
\def\delfiyu{\Delta \bfe (y,y_i^u; \bx_i^u)}
\def\liy{L(y,y_i)}
\def\liys{L(y,y_i^s)}
\def\liyu{L(y,y_i^u)}
\def\bxbar{\bar{\bx}}
\def\ybar{\bar{y}}
\def\ibar{\bar{i}}
\def\delc{\delta c}
\def\to{{\rightarrow}}

\def\delfiybaryi{\Delta \bf_i (\bar{\by}_i)}

\def\Fiy{F_i ({\by})}
\def\Alpha{{\mbox{\boldmath $\alpha$}}}
\def\Alphai{{\mbox{\boldmath $\alpha_i$}}}
\def\delAlphai{{\mbox{\boldmath $\delta \alpha_i$}}}
\def\delalphaiy{{\mbox{\boldmath $\delta \alpha_i({\by})$}}}

\def\ool{\frac{1}{l}}
\def\oon{\frac{1}{n}}

\def\Y{\mathscr Y}
\def\X{\mathscr X}

\def\liybari{l_i(\bar{\by}_i)}

\def\dow{\partial}

\newpage

\section{Introduction}
\label{intro}
%\vspace*{-0.1in}

Consider the following supervised learning problem corresponding to a general structured output prediction problem:
\begin{eqnarray}
\min_{\bw, \xi^s} \dsp F^s(\bw) =  \dsp  \half {\| {\bw} \|}^2 + \ool \sum_{i=1}^l \xi_i^s
\label{eq:sup}
\end{eqnarray}
where $\xi_i^s=\xi(\bw,\bx_i^s,y_i^s)$ is the loss term and $\{(\bx_i^s,y_i^s)\}_{i=1}^l$ is the set of labeled examples. For example, in large margin and maxent models we have
\begin{eqnarray}
%\xi(\bw,\bx_i,y_i) = \max_y  \liy - {\bw}^T {\delfiy}, \dsp\dsp \delfiy = \bfe(y_i;\bx_i) - \bfe(y;\bx_i) \label{eq:lmloss} \\
\xi(\bw,\bx_i,y_i) = \max_y  \liy - {\bw}^T {\delfiy} \label{eq:lmloss} \\
\xi(\bw,\bx_i,y_i) = - {\bw}^T \bfe(y_i;\bx_i) + \log Z \label{eq:mentloss}
\label{eq:loss}
\end{eqnarray}
where $\delfiy = \bfe(y_i;\bx_i) - \bfe(y;\bx_i)$ and $Z = \sum_y \exp({\bw}^T \bfe(y;\bx_i))$. Text classification problems involve a rich and large feature space (e.g., bag-of-words features) and so linear classifiers work very well~\cite{Joachims1999}.
%In this paper [ssrajan]
We particularly focus on multi-class and hierarchical classification problems (and hence our use of scalar notation for $y$).
In multi-class problems $y$ runs over the classes and, ${\bw}$ and $\bfe(y;\bx_i)$ have one component for each class, with the component corresponding to $y$ turned on.
More generally, in hierarchical classification problems, $y$ runs over the set of leaf nodes of the hierarchy and, ${\bw}$ and $\bfe(y;\bx_i)$ consist of one component for each node of the hierarchy, with the node components in the path to leaf node $y$ turned on.
$\lambda > 0$ is a regularization parameter. A good default value for $\lambda$ can be chosen depending on the loss function used.\footnote{In the experiments of this paper, for multi-class and hierarchical classification with large margin loss, we use $\lambda=10$ and, for binary maxent loss we use $\lambda=10^{-3}$.}
The superscript $s$ denotes `supervised'; we will use superscript $u$ to denote elements corresponding to unlabeled examples.

In semi-supervised learning we use a set of unlabeled examples, $\{\bx_i^u\}_{i=1}^n$ and include the determination of the labels of these examples as part of the training process:
\begin{eqnarray}
\min_{\bw, \by^u} \dsp F^s(\bw) + \frac{C^u}{n} \sum_{i=1}^n \xi_i^u  \label{eq:semisup1}  \\
{\rm s.t.} \dsp \sum_{i=1}^n \delta(y,y_i^u) = n(y) \dsp \forall y \label{eq:semisup2}
\end{eqnarray}
where $\by^u = \{y_i^u\}$, $\xi_i^u = \xi(\bw,\bx_i^u,y_i^u)$  and $\delta$ is the Kronecker delta function.
$C^u$ is a regularization parameter for the unlabeled part. A good default value is $C^u=1$; we use this value in all our experiments.
\eqref{eq:semisup2} consists of constraints on the label counts that come from domain knowledge.
(In practice, one specifies $\phi(y)$, the fraction of examples in class $y$; then the values in $\{\phi(y)n\}$ are rounded to integers $\{n(y)\}$ in a suitable way so that $\sum_y n(y)=n$.\footnote{We will assume that quite precise values are given for $\{n(y)\}$. The effect of noise in these values on the semi-supervised solution needs a separate study.}) Such constraints are crucial for the effective solution of the semi-supervised learning problem; without them the semi-supervised solution tends to move towards assigning the majority class label to most unlabeled examples. In more general structured prediction problems~\eqref{eq:semisup2} may include other domain constraints~\cite{Chang2007}. In this paper we will use just the label constraints in~\eqref{eq:semisup2}.

Inspired by the effectiveness of the TSVM model of \shortcite{Joachims1999}, there have been a number of works on the solution of~ \eqref{eq:semisup1}-\eqref{eq:semisup2} for binary classification with large margin losses. These methods fall into one of two types: (a) combinatorial optimization; and (b) continuous optimization. See~\cite{Chapelle2008,Chapelle2006} for a detailed coverage of various specific methods falling into these two types.
In combinatorial optimization the label set $\by^u$ is determined together with $\bw$. It is usual to use a sequence of alternating optimization steps (fix $\by^u$ and solve for $\bw$, and then fix $\bw$ and solve for $\by^u$) to obtain the solution. An important advantage of doing this is that each of the sub-optimization problems can be solved using simple and/or standard solvers.
In continuous optimization $\by^u$ is eliminated and the resulting (non-convex) optimization problem is solved for $\bw$ by minimizing
%the objective function,
\begin{eqnarray}
%F^s(\bw) + \frac{C^u}{n} \sum_{i=1}^n \rho(\bw,\bx_i^u) \dsp {\rm where} \rho(\bw,\bx_i^u) = \min_{y^u} \xi(\bw,\bx_i^u,y_i^u)  \label{eq:semisup3}
F^s(\bw) + \frac{C^u}{n} \sum_{i=1}^n \rho(\bw,\bx_i^u)  \label{eq:semisup3}
\end{eqnarray}
where $\rho(\bw,\bx_i^u) = \min_{y^u} \xi(\bw,\bx_i^u,y_i^u)$. The loss function $\xi$ as well as $\rho$ are usually smoothed so that the objective function is differentiable and gradient-based optimization techniques can be employed. Further, the constraints in \eqref{eq:semisup2} involving $\by^u$ are replaced by smooth constraints on $\bw$ expressing balance of the mean outputs of each label over the labeled and unlabeled sets.

\shortcite{Zien2007} extended the continuous optimization approach to~\eqref{eq:semisup3} for multi-class and structured output problems. But their experiments only showed limited improvement over supervised learning. The combinatorial optimization approach, on the other hand, has not been carefully explored beyond binary classification.
Methods based on semi-definite programming~\cite{Xu2006,Debie2004} are impractical, even for medium size problems.
One-versus-rest and one-versus-one ideas have been tried, but it is unclear if they work well:~\shortcite{Zien2007} and ~\shortcite{Zubiaga2009} report failure while~\shortcite{Bruzzone2006} use a heuristic implementation and report success in one application domain. Unlike these methods which have binary TSVM as the basis, we take up an implementation of the approach for the direct multi-class and hierarchical classification formulation in~\eqref{eq:semisup1}-\eqref{eq:semisup2}. The special structure in constraints allows the $\by^u$ determination step to reduce to a degenerate transportation linear programming problem. So the well-known transportation simplex method can be used to obtain $\by^u$. We show that even this method is not efficient enough. As an alternative we suggest an effective and much more efficient heuristic label switching algorithm. For binary classification problems this algorithm is an improved version of the multiple switching algorithm developed by~\shortcite{Sindhwani2006} for TSVM.
Experiments on a number of multi-class and hierarchical classification datasets show that, like the TSVM method of binary classification, our method yields a strong lift in performance over supervised learning, especially when the number of labeled examples is not sufficiently large.

The applicability of our approach to general loss functions is a key advantage.
Specialized to maxent losses, the method offers an interesting alternative to the idea of entropy regularization~\cite{Grandvalet2004} and related methods~\cite{Lee2006}. For maxent losses, there also exist other methods such as expectation regularization/constraint~\cite{Mann2010} and posterior regularization~\cite{Gartner2005,Graca2007,Ganchev2009} which use unlabeled examples only to enforce the constraints in~\eqref{eq:semisup2}. In section~\ref{maxent} we compare our approach with these methods on binary classification and point out that our method gives a stronger performance.

\section{Semi-Supervised Learning Algorithm}
\label{semisupalgm}
%\vspace*{-0.1in}

The semi-supervised learning algorithm for multi-class and hierarchical classification problems follows the spirit of the TSVM algorithm~\cite{Joachims1999}. Algorithm 1 gives the steps. It consists of an initialization part (steps 1-9) that sets starting values for $\bw$ and $\by^u$, followed by an iterative part (steps 10-15) where $\bw$ and $\by^u$ are refined by semi-supervised learning. Using exactly the same arguments as those in~\cite{Joachims1999,Sindhwani2006} it can be proved that Algorithm 1 is convergent.

Initialization of $\bw$ is done by solving the supervised learning problem. This $\bw$ can be used to predict $\by^u$. However such a $\by^u$ usually violates the constraints in~\eqref{eq:semisup2}. To choose a $\by^u$ that satisfies~\eqref{eq:semisup2}, we do a greedy modification of the predicted $\by^u$. Steps 3-9 of Algorithm 1 give the details.

The iterative part of the algorithm consists of an outer loop and an inner loop. In the outer loop (steps 10-15) the regularization parameter $C^u$ is varied from a small value to the final value of $1$ in annealing steps. This is done to avoid drastic switchings of the labels in $\by^u$, which helps the algorithm reach a better minimum of~\eqref{eq:semisup1}-\eqref{eq:semisup2} and hence achieve better performance. For example, on ten runs of the multi-class dataset, {\it 20NG} (see Table 1) with 100 labeled examples and $10,000$ unlabeled examples, the average macro F values on test data achieved by supervised learning, Algorithm 1 without annealing and Algorithm 1 with annealing are, respectively, 0.4577, 0.5377 and 0.6253. Similar performance differences are seen on other datasets too.

The inner loop (steps 11-14) does alternating optimization of $\bw$ and $\by^u$ for a given $C^u$. In steps 12 and 13 we use the most recent $\bw$ and $\by^u$ as the starting points for the respective sub-optimization problems. Because of this, the overall algorithm remains very efficient in spite of the many annealing steps involving $C^u$. Typically, the overall cost of the algorithm is only about 3-5 times that of solving a supervised learning problem involving $(n+l)$ examples. For step 12 one can employ any standard algorithm suited to the chosen loss function. In the rest of the section we will focus on step 13.

\begin{algorithm}
	\caption{\textit{Semi-Supervised Learning Algorithm}}
	\label{tsvm}
 	\begin{algorithmic}[1]
 	\renewcommand{\algorithmicrequire}{\textbf{Input:}}
	\STATE Solve the supervised learning problem, \eqref{eq:sup} and get $\bw$.
    \STATE Set initial labels for unlabeled examples, $\by^u$ using steps 3-9 below.
    \STATE Set $Y=\{y\}$, the set of all classes, $A_y=\emptyset\dsp\forall y$, and $I=\{1,\ldots,n\}$.
    \REPEAT
        \STATE $S_i=\max_{y\in Y} \bw^T \bfe(y;\bx_i^u)$ and $y_i=\arg\max_{y\in Y} \bw^T \bfe(y;\bx_i^u)\dsp  \forall i\in I$.
        \STATE Sort $I$ by decreasing order of $S_i$.
        \STATE By order allocate $i$ to $A_{y_i}$ while not exceeding sizes specified by $n(y_i)$.
        \STATE Remove all allocated $i$ from $I$ and remove all saturated $y$ (i.e., $|A_y|=n(y)$) from $Y$.
	\UNTIL {$Y=\emptyset$}
    \FOR {$C^u = \{ 10^{-4}, 3\times 10^{-4}, 10^{-3}, 3\times 10^{-3},\ldots,1 \} $ (in that order)}
        \REPEAT
            \STATE Solve \eqref{eq:semisup1} for $\bw$ with $\by^u$ fixed.
            \STATE Solve \eqref{eq:semisup1}-\eqref{eq:semisup2} for $\by^u$ with $\bw$ fixed.
        \UNTIL step 13 does not alter $\by^u$
    \ENDFOR
  \end{algorithmic}
\end{algorithm}

\subsection{Linear programming formulation}
Let us now consider optimizing $\by^u$ with fixed $\bw$. Let us represent each $y_i^u$ in a 1-of-$m$ representation by defining boolean variables $z_{iy}$ and requiring that, for each $i$, exactly one $z_{iy}$ takes the value 1. This can be done by using the constraint $\sum_y z_{iy}=1$ for all $i$. The label constraints become $\sum_i z_{iy}=n(y)$ for all $y$. Let $c_{iy}=\xi(\bw,\bx_i^u,y)$. With these definitions the optimization problem of step 13 becomes (irrespective of the type of loss function used) the integer linear programming problem,
\begin{eqnarray}
\min \dsp \sum_{i,y} c_{iy} z_{iy} \dsp {\rm s.t.} \label{eq:lp} \\
\sum_y z_{iy} = 1 \dsp \forall i, \dsp \sum_i z_{iy} = n(y) \dsp \forall y, \label{eq:lpcons} \\
z_{iy} \in \{0,1\} \dsp \forall i, y \label{eq:ic}
\end{eqnarray}
This is a special case of the well known Transportation problem~\cite{Hadley1963} in which the constraint matrix satisfies unimodularity conditions; hence, the solution of the integer linear programming problem \eqref{eq:lp}-\eqref{eq:ic} is same as the solution of the linear programming (LP) problem,~\eqref{eq:lp}-\eqref{eq:lpcons} (note: in LP the integer constraints are left out), i.e., at LP optimality~\eqref{eq:ic} holds automatically. Previous works~\cite{Joachims1999,Sindhwani2006} do not make this neat connection to linear programming.
The constraints $\sum_y z_{iy}=1\dsp\forall i$ allow exactly $n$ non-zero elements in $\{z_{iy}\}_{iy}$; thus
there is degeneracy of order $m$, i.e., there are $(n+m)$ constraints but only $n$ non-zero solution elements.

\subsection{Transportation simplex method}
The transportation simplex method (a.k.a., stepping stone method)~\cite{Hadley1963} is a standard and generally efficient way of solving LPs such as~\eqref{eq:lp}. However, it is not efficient enough for typical large scale learning situations in which $n$, the number of unlabeled examples is large and $m$, the number of classes, is small. Let us see why. Each iteration of this method starts with a basis set of $n+m-1$ basis elements. Then it computes reduced costs for all remaining elements. This step requires $O(nm)$ effort. If all reduced costs are non-negative then it implies that the current solution is optimal. If this condition does not hold, elements which have negative reduced costs are potential elements for entering the basis.\footnote{Presence of negative reduced costs may not mean that the current solution is non-optimal. This is due to degeneracy. It is usually the case that, even when an optimal solution is reached, the transportation algorithm requires several end steps to move the basis elements around to reach an end state where positive reduced costs are seen.} One non-basis element with a negative reduced cost (say, the element with the most negative reduced cost) is chosen. The algorithm now moves the solution to a new basis in which an element of the previous basis is replaced by the newly entering element. This operation corresponds to moving a chosen set of examples between classes in a loop so that the label constraints are not violated. The number of such iterations is observed to be $O(nm)$ and so, the algorithm requires $O(n^2m^2)$ time. Since $n$ can be large in semi-supervised learning, the transportation simplex algorithm is not sufficiently efficient. The main cause of inefficiency is that the step (one basis element changed) is too small for the amount of work put in (computing all reduced costs)!
\begin{algorithm}
	\caption{\textit{Switching Algorithm to solve~\eqref{eq:lp}-\eqref{eq:ic} }}
	\label{switch}
 	\begin{algorithmic}[1]
 	\renewcommand{\algorithmicrequire}{\textbf{Input:}}
 	%\renewcommand{\algorithmicreturn}{\textbf{return}}
	%\STATE Set {\it done=false}
    \REPEAT
      	\FOR{each class pair $(y,\ybar)$}
			\STATE Compute $\delc(i,y,\ybar)$ for all $i$ in class $y$ and sort
            the elements in increasing order of $\delc$ values.
            \STATE Compute $\delc(\ibar,\ybar,y)$ for all $\ibar$ in class $\ybar$ and sort the elements in increasing order of $\delc$ values.
			\STATE Align these two lists (so that the best pair is at the top)
            to form a switch list of 5-tuples, $\{(i,y,\ibar,\ybar,\rho(i,y,\ibar,\ybar)\}$.
            \STATE Remove any 5-tuple with $\rho(i,y,\ibar,\ybar)\ge 0$.
		\ENDFOR
		\STATE Merge all the switch lists into one and sort the 5-tuples by increasing
        order of $\rho$ values.
        \WHILE{switch list is non-empty}
            \STATE Pick the top 5-tuple from the switch list; let's say it is $(i,y,\ibar,\ybar,\rho(i,y,\ibar,\ybar))$. Move $i$ to class $\ybar$ and move $\ibar$ to class $y$.
            \STATE From the remaining switch list remove all 5-tuples involving either $i$ or $\ibar$.
        \ENDWHILE
	\UNTIL {the merged switch list from step 8 is empty}
  \end{algorithmic}
\end{algorithm}

\subsection{Switching algorithm}
We now propose an efficient heuristic {\it switching algorithm} for solving~\eqref{eq:lp}-\eqref{eq:ic} that is suited to the case where $n$ is large but $m$ is small. The main idea is to use only pairwise switching of labels between classes in order to improve the objective function. (Note that switching makes sure that the label constraints are not violated.) This algorithm is sub-optimal for $m\ge 3$, but still quite powerful because of two reasons: (a) the solution obtained by the algorithm is usually close to the true optimal solution; and (b) reaching optimality precisely is not crucial for the alternating optimization approach (steps 12 and 13 of Algorithm 1) to be effective.

Let us now give the details of the switching algorithm. Suppose, in the current solution, example $i$ is in class $y$. Let us say we move this example to class $\ybar$. The change in objective function due to the move is given by $\delc(i,y,\ybar) = c_{i\ybar}-c_{iy}$. Suppose we have another example $\ibar$ which is currently in class $\ybar$ and we switch $i$ and $\ibar$, i.e., move $i$ to class $\ybar$ and move $\ibar$ to class $y$. The resulting change in objective function is given by
\begin{equation}
\rho(i,y,\ibar,\ybar) = \delc (i,y,\ybar) + \delc (\ibar,\ybar,y)
\end{equation}
The more negative $\rho(i,y,\ibar,\ybar)$ is, the better will be the objective function reduction due to the switching of $i$ and $\ibar$. The algorithm looks greedily for finding as many good switches as possible at a time. Algorithm 2 gives the details. Steps 2-12 consist of one major greedy iteration and has cost $O(nm^2)$. Steps 2-7 consist of the background work needed to do the greedy switching of several pairs of examples in steps 9-12. Step 11 is included because, when $i$ and $\ibar$ are switched, data related to any 5-tuple in the remaining switch list that involves either $i$ or $\ibar$ is messed up. Removing such elements from the remaining switched list allows the algorithm to continue finding more pairs to apply switching without a need for repeating steps 2-7. It is this multiple switching idea that gives the needed efficiency lift over the transportation simplex algorithm.

The algorithm is convergent due to the following reasons: the algorithm only performs switchings which reduce the objective function; thus, once a pair of examples is switched, that pair will not be switched again; and, the number of possible switchings is finite. A typical run of Algorithm 2 requires about 3 loops through steps 2-12. Since this algorithm only allows pairwise switching of examples, it cannot assure that the class assignments resulting from it will be optimal for~\eqref{eq:lp}-\eqref{eq:ic} if $m\ge 3$. However, in practice the objective function achieved by the algorithm is very close to the true optimal value; also, as pointed out earlier, reaching true optimality turns out to be not crucial for good performance of the semi-supervised algorithm.

\subsection{Comparison of the algorithms}
Figure 1 shows the performance of transportation simplex
%\begin{wrapfigure}{l}{0.5\textwidth}
and switching algorithms on the {\it Ohscal} dataset~\cite{Forman2003} with 100/5581 labeled/unlabeled examples.
Note that the cpu times ($x$-axis) are in log scale. While transportation simplex requires $100$ secs, the switch algorithm reaches close to optimal well within a second. On the binary classification dataset, {\it aut-avn}~\cite{Sindhwani2006} with  100/35888 labeled/unlabeled examples, the switch algorithm reaches exact optimality requiring only $0.1$ seconds while transportation simplex requires $30$ minutes!

If $m$ is large then steps 2-7 of Algorithm 2 can become expensive. We have applied the switching algorithm to datasets that have $m\le 105$, but haven't observed any inefficiency. If $m$ happens to be much larger then steps 2-7 can be modified to work with a suitably chosen subset of class pairs instead of all possible pairs.

\begin{figure}
  %\label{fig:transp}
  \begin{center}
    \vspace*{-1.2in}
    %\vspace*{-1.4in}
    %\hspace*{-0.3in}
    %\includegraphics[width=0.48\textwidth]{plots/SpeedComparison.pdf}
    \includegraphics[width=0.6\textwidth]{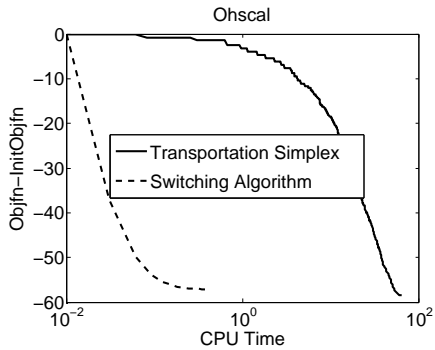}
  \end{center}
  %\vspace*{-1.4in}
  \vspace*{-1.5in}
  \caption{{Comparison of costs of Transportation simplex and Switching algorithms on {\it Ohscal} dataset with 100/5581 labeled/unlabeled examples, on the first entry to step 13 of Algorithm 1. The vertical axis gives the change in objective function from the initial value.}}
%\end{wrapfigure}
\end{figure}

\subsection{Relation with binary TSVM methods}
Consider the case $m=2$ (binary classification). There is only a single class pair and so step 11 is not needed. Joachims' original TSVM method~\cite{Joachims1999} corresponds to the version of Algorithm 2 in which only one switch (the top candidate in step 10) is made. Sindhwani and Keerthi's multiple switching algorithm~\cite{Sindhwani2006} is more efficient than Joachims' method and corresponds to doing one outer loop of Algorithm 2, i.e., steps 2-12.
%Neither of these methods ensures reaching optimality.
Algorithm 2 is more improved and is also optimal for $m=2$. This can be proved by noting the following: the algorithm is convergent; at convergence there is no switching pair which improves the objective function; and, for $m=2$ a transportation simplex step corresponds to switching labels for a set of example pairs. Thus, if the convergent solution is not optimal, a transportation simplex iteration can be applied to find at least one switching pair that leads to objective function reduction, which is a contradiction.

\section{Experiments with large margin loss}
\label{expts}
%\vspace*{-0.1in}

In this section we give results of experiments on our method as applied to multi-class and hierarchical classification problems using the large margin loss function, \eqref{eq:lmloss}. We used the loss, $L(y,y_i)=\delta(y,y_i)$. Eight multi-class datasets and two hierarchical classification datasets were used. Properties of these datasets~\cite{Lang1995,Forman2003,McCallum1998,Lewis2004,LeCun,Tibshirani} are given in Table~1. Most of these datasets are standard text classification benchmarks. We include two image datasets, {\it mnist} and {\it usps} to point out that our methods are useful in other application domains too. {\it rcv-mcat} is a subset of rcv1~\cite{Lewis2004} corresponding to the sub-tree belonging to the high level category {\sc MCAT} with seven leaf nodes consisting of the categories, {\sc Equity}, {\sc Bond}, {\sc Forex}, {\sc Commodity}, {\sc Soft}, {\sc Metal} and {\sc Energy}.
In one run of each dataset, 50\% of the examples were randomly chosen to form the unlabeled set, $U$; 20\% of the examples were put aside in a set $L$ to form labeled data; the remaining data formed the test set. Ten such runs were done to compute the mean and standard deviation of (test) performance. Performance was measured in terms of Macro F (mean of the F values associated with various classes).

\begin{table*}
\begin{small}
%\label{table:dset}
%\vspace*{-0.2in}
\centering
\caption{Properties of datasets. $N$ : number of examples, $d$ : number of features, $m$ : number of classes, Type: M=Multi-Class; H=Hierarchical, with D=Depth and I=\# Internal Nodes}
\vspace{0.2in}
\begin{tabular}{|c|c|c|c|c|c|c|c|c|c|c|} \hline
         & {\it 20NG}  & {\it la1}   & {\it webkb} & {\it ohscal} & {\it reut8}  & {\it sector} & {\it mnist} & {\it usps} & {\it 20NG}    & {\it rcv-mcat}  \\ \hline
$N$      & 19928 & 3204  & 8277  & 11162  & 8201   & 9619   & 70000 & 9298 & 19928   & 154706    \\ \hline
$d$      & 62061 & 31472 & 3000  & 11465  & 10783  & 55197  & 779   & 256  & 62061   & 11429     \\ \hline
$m$      & 20    & 6     & 7     & 10     & 8      & 105    & 10    & 10   & 20      & 7         \\ \hline
Type     & M     & M     & M     & M      & M      & M      & M     & M    & H       & H         \\
D/I      &       &       &       &        &        &        &       &      & 3/8     & 2/10      \\ \hline
\end{tabular}
%\vspace*{0.3in}
\end{small}
\end{table*}
\begin{figure*}
%\label{fig:tax}
\begin{center}
%\framebox[4.0in]{$\;$}
%\fbox{\rule[-.5cm]{0cm}{4cm} \rule[-.5cm]{4cm}{0cm}}
\vspace*{-0.2in}
   \includegraphics[width=0.45\linewidth]{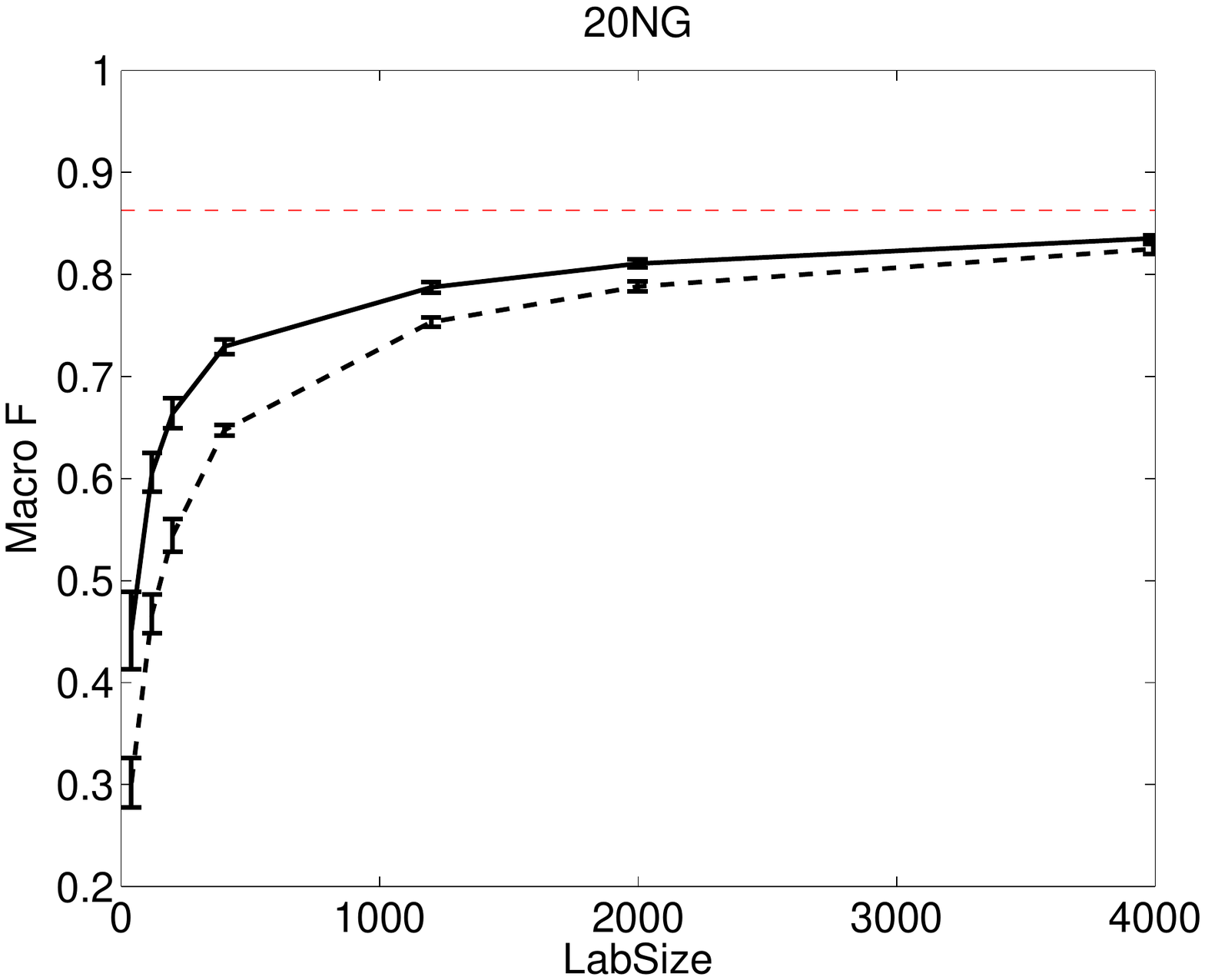}
   %\hspace{1cm}
   \includegraphics[width=0.45\linewidth]{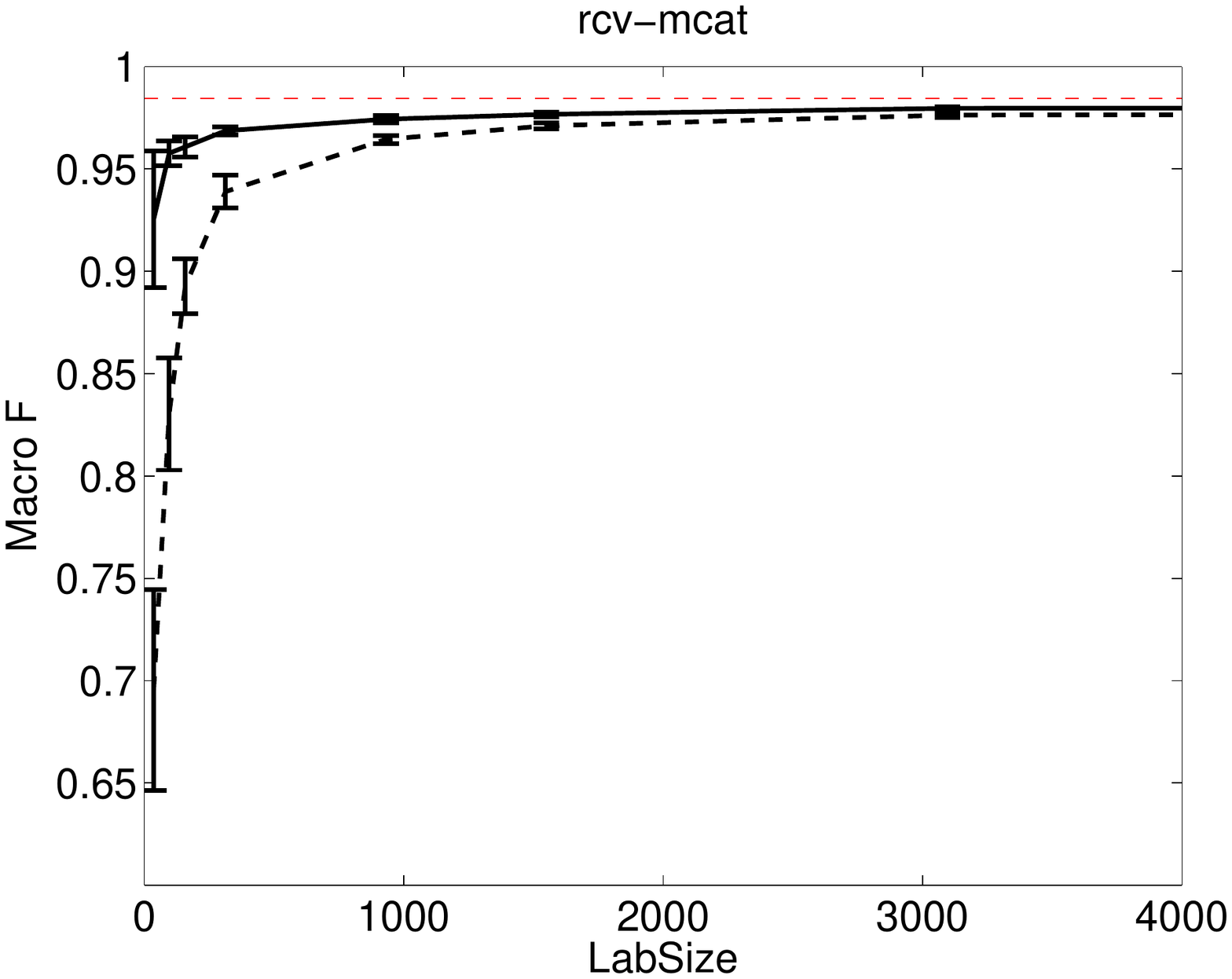} \\
\end{center}
\vspace*{-0.9in}
\caption{{Hierarchical classification datasets: Variation of performance (Macro F) as a function of the number of labeled examples (LabSize). Dashed black line corresponds to supervised learning; Continuous black line corresponds to the semi-supervised method; Dashed horizontal red line corresponds to the supervised classifier built using $L$ and $U$ with their labels known.}}
\end{figure*}
%\begin{wrapfigure}{l}{0.5\textwidth}
In the first experiment, we fixed the number of labeled examples (to 80) and varied the number of unlabeled examples from small to big values. The variation of performance as a function of the number of unlabeled examples, for the multi-class dataset, {\it 20NG}, is given in Figure 2. Performance steadily improves as more unlabeled data is added. The same holds in other datasets too.

\begin{figure}
  %\label{fig:transp}
  \begin{center}
    %\includegraphics[width=0.48\textwidth]{gull}
    %\vspace*{-0.8in}
    \vspace*{-1.0in}
    %\hspace*{-0.3in}
    %\includegraphics[width=0.48\textwidth]{vary_unlab_plots/Lcurve_F_20NG_uv.pdf}
    \includegraphics[width=0.6\textwidth]{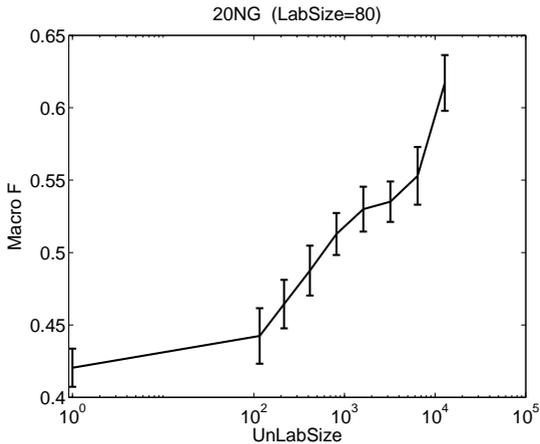}
  \end{center}
  \vspace*{-1.0in}
  %\vspace*{-0.8in}
  \caption{{{\it 20NG:} Variation of performance (Macro F) as a function of the number of unlabeled examples (UnLabSize), with the number of labeled examples fixed at 80.}}
%\end{wrapfigure}
\end{figure}
Next we fixed the unlabeled data to $U$ and varied the labeled data size from small values up to $|L|$. This is an important study for semi-supervised learning methods since their main value is when labeled data is sparse (lower side of the learning curve). The variation of performance as a function of the number of unlabeled examples is shown for the two hierarchical classification datasets in Figure 3 and, results for six multi-class datasets in Figure 4. We observed that the performance on the $20NG$ dataset was almost same in the multi-class and hierarchical classification scenarios. Also, the performance was similar on the $MNIST$ and $USPS$ datasets. Clearly, semi-supervised learning is very useful and yields good improvement over supervised learning especially when labeled data is sparse. The degree of improvement is sharp in some datasets (e.g., {\it reut8}) and mild in some datasets (e.g., {\it sector}).

While the semi-supervised method is successful in linear classifier settings such as in text classification and natural language processing, we want to caution, like~\cite{Chapelle2008}, that it may not work well on datasets originating from nonlinear manifold structure.

\begin{figure*}
\begin{center}
%\framebox[4.0in]{$\;$}
%\fbox{\rule[-.5cm]{0cm}{4cm} \rule[-.5cm]{4cm}{0cm}}
\vspace*{-0.9in}
   \begin{comment}
   \includegraphics[width=0.35\linewidth]{plots/Lcurve_F_20NG.pdf}
   \hspace{1cm}
   \includegraphics[width=0.35\linewidth]{plots/Lcurve_F_mnist.pdf} \\
   \vspace*{-1.45in}
   \end{comment}
   \includegraphics[width=0.45\linewidth]{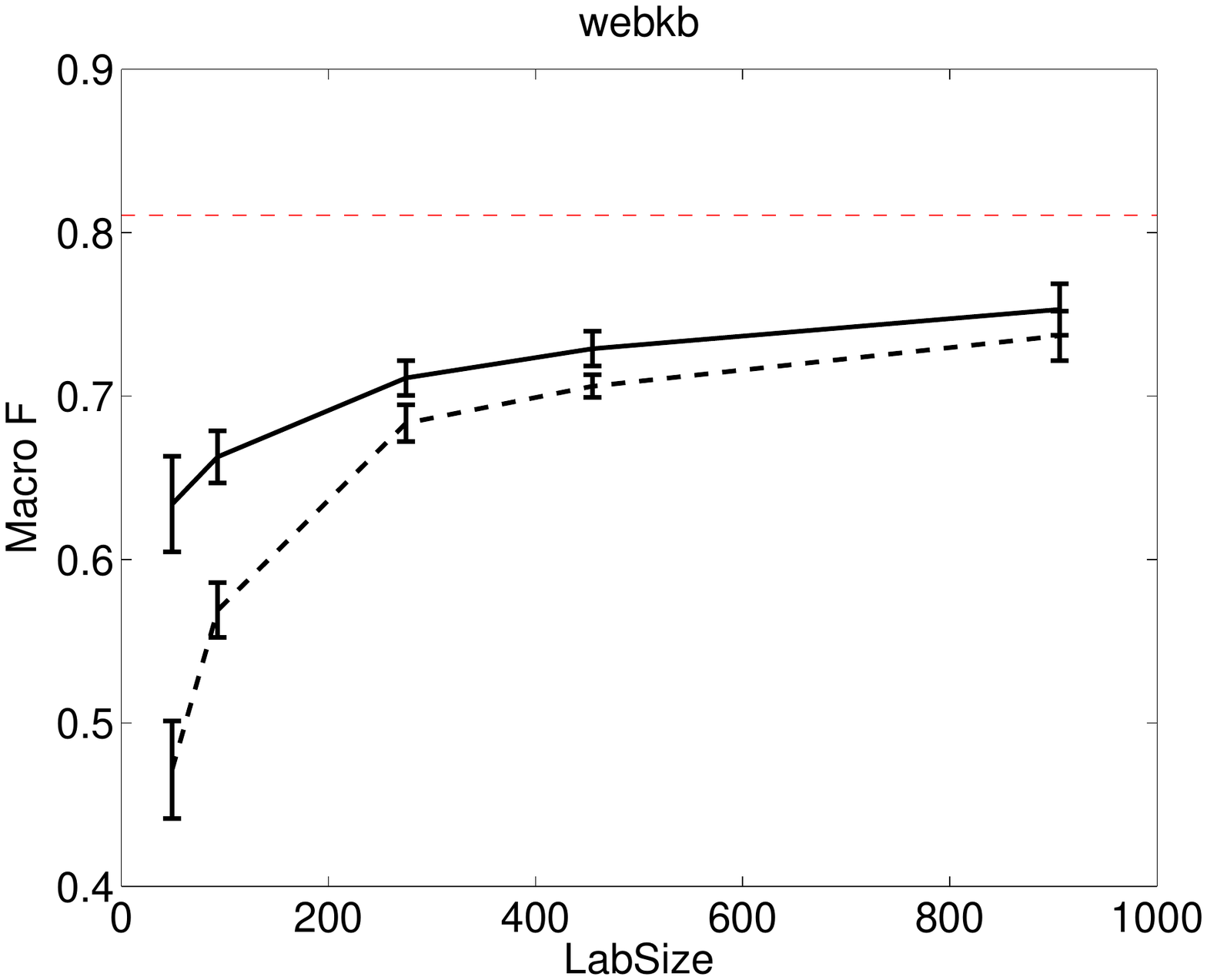}
   %\hspace{1cm}
   \includegraphics[width=0.45\linewidth]{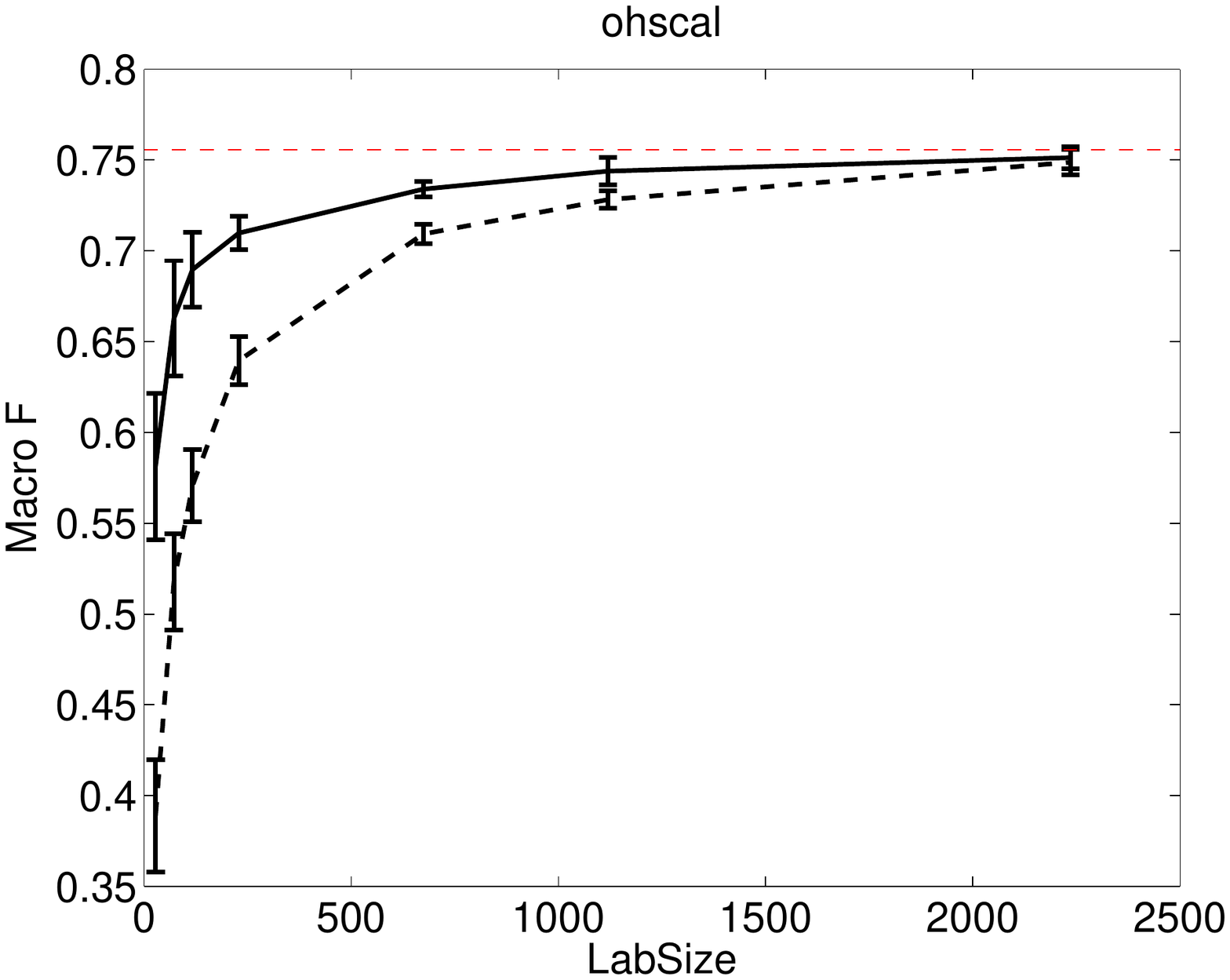} \\
   \vspace*{-1.3in}
   \includegraphics[width=0.45\linewidth]{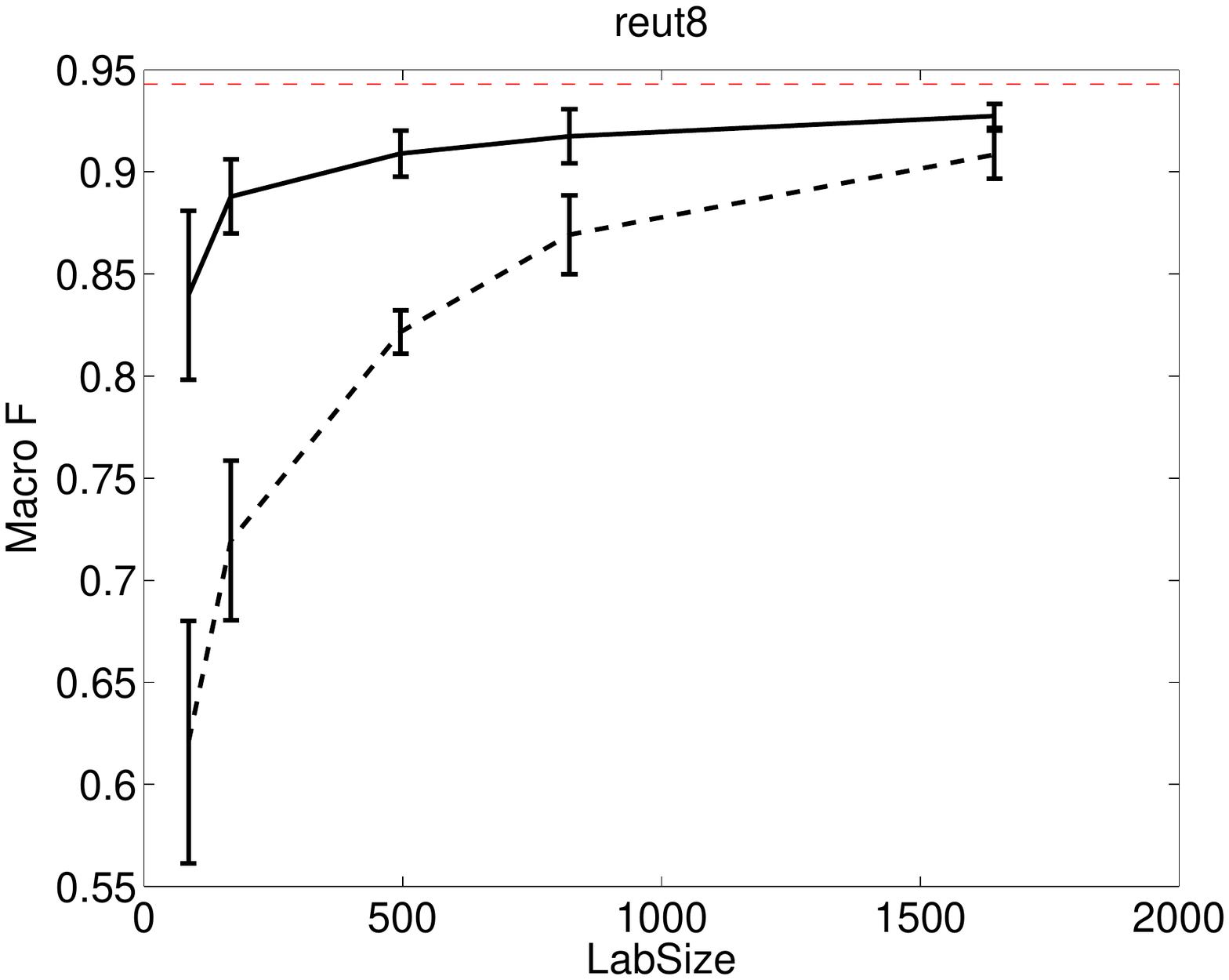}
   %\hspace{1cm}
   \includegraphics[width=0.45\linewidth]{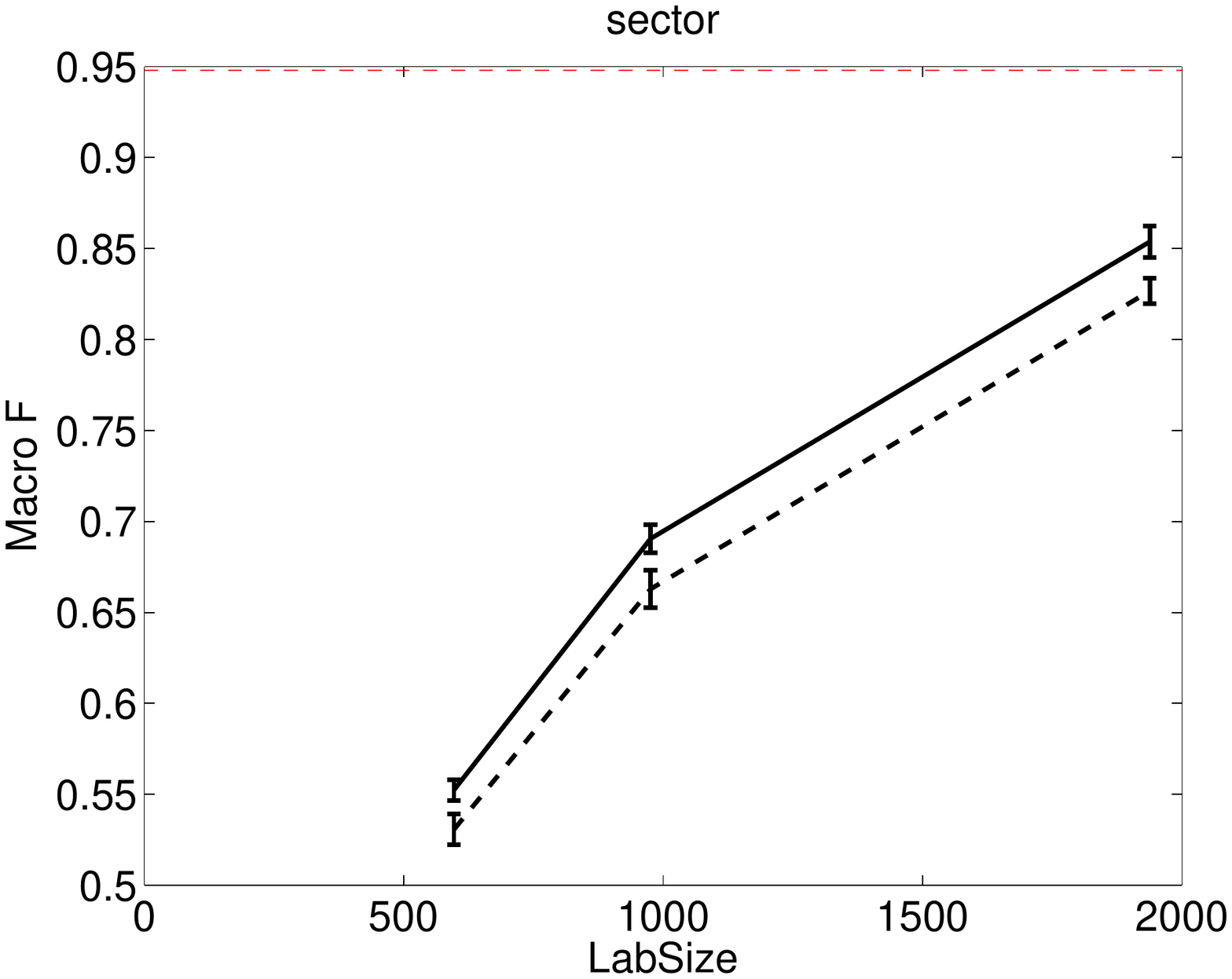} \\
   %\begin{comment}
   \vspace*{-1.3in}
   \includegraphics[width=0.45\linewidth]{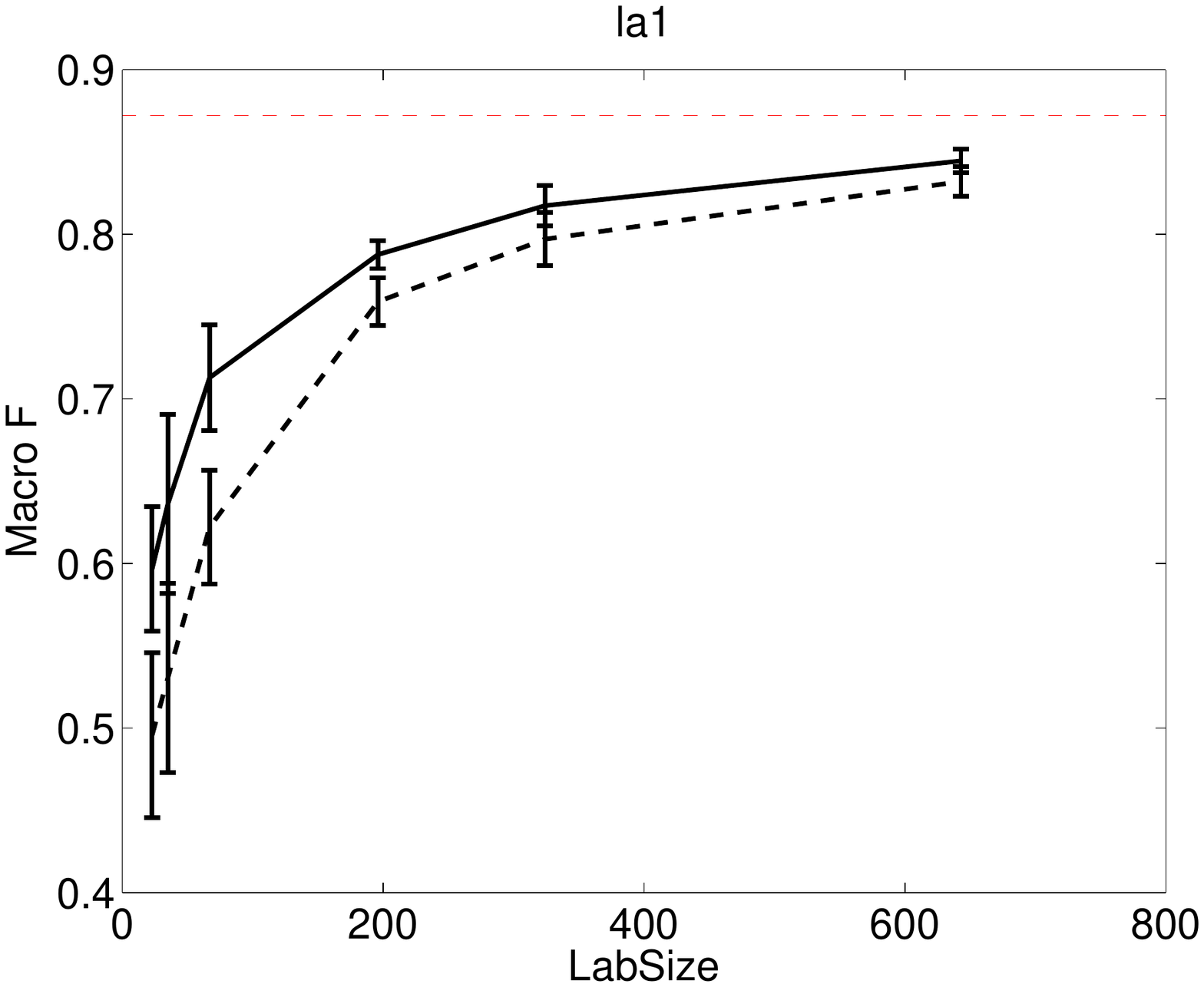}
   %\hspace{1cm}
   \includegraphics[width=0.45\linewidth]{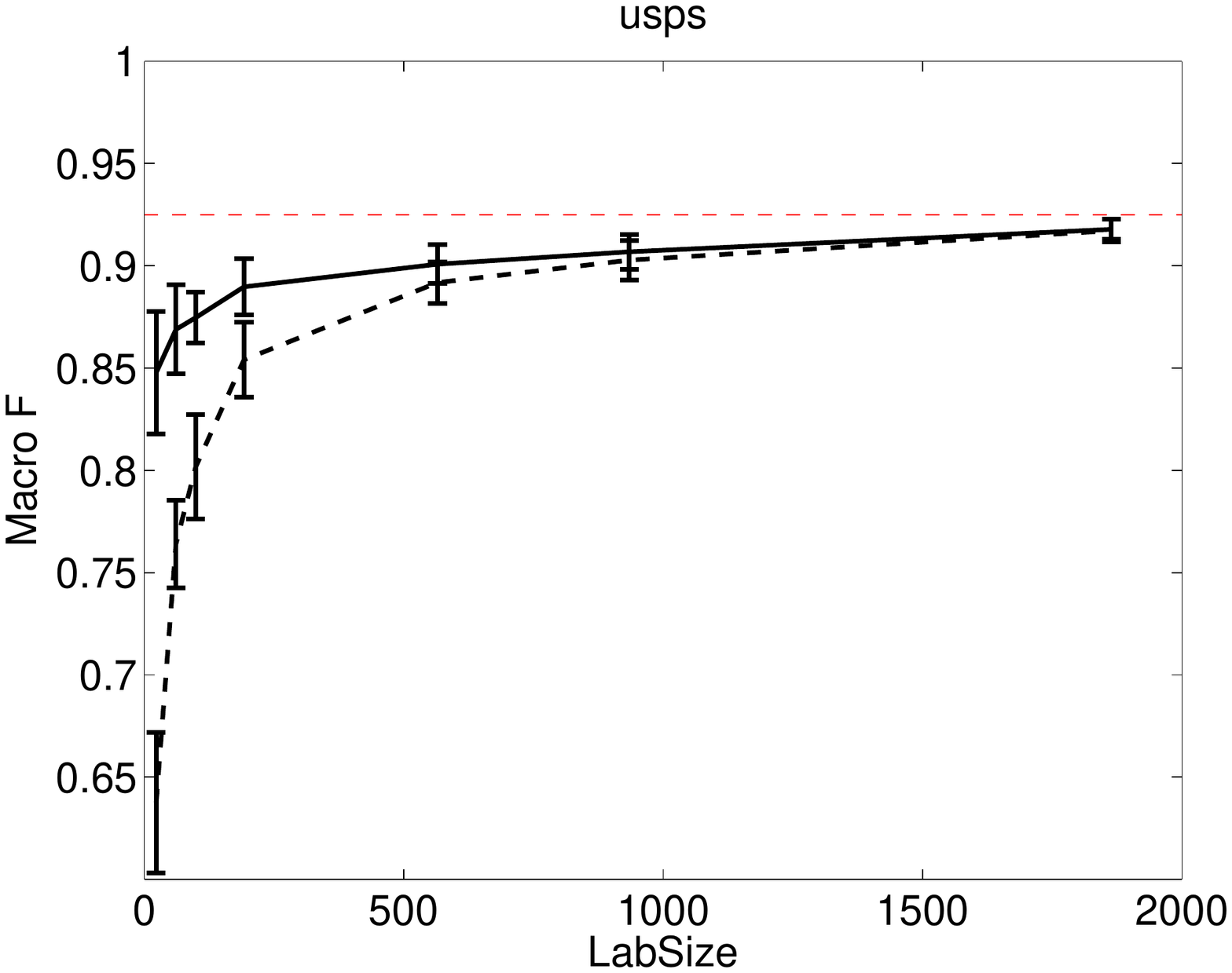}
   %\end{comment}
\end{center}
\vspace*{-0.8in}
\caption{{Multi-class datasets: Variation of performance (Macro F) as a function of the number of labeled examples (LabSize). Dashed black line corresponds to supervised learning; Continuous black line corresponds to the semi-supervised method; Dashed horizontal red line corresponds to the supervised classifier built using $L$ and $U$ with their labels known.}}
\end{figure*}

\begin{figure*}
\begin{center}
%\framebox[4.0in]{$\;$}
%\fbox{\rule[-.5cm]{0cm}{4cm} \rule[-.5cm]{4cm}{0cm}}
%\vspace*{-1.0in}
   \includegraphics[width=0.45\linewidth]{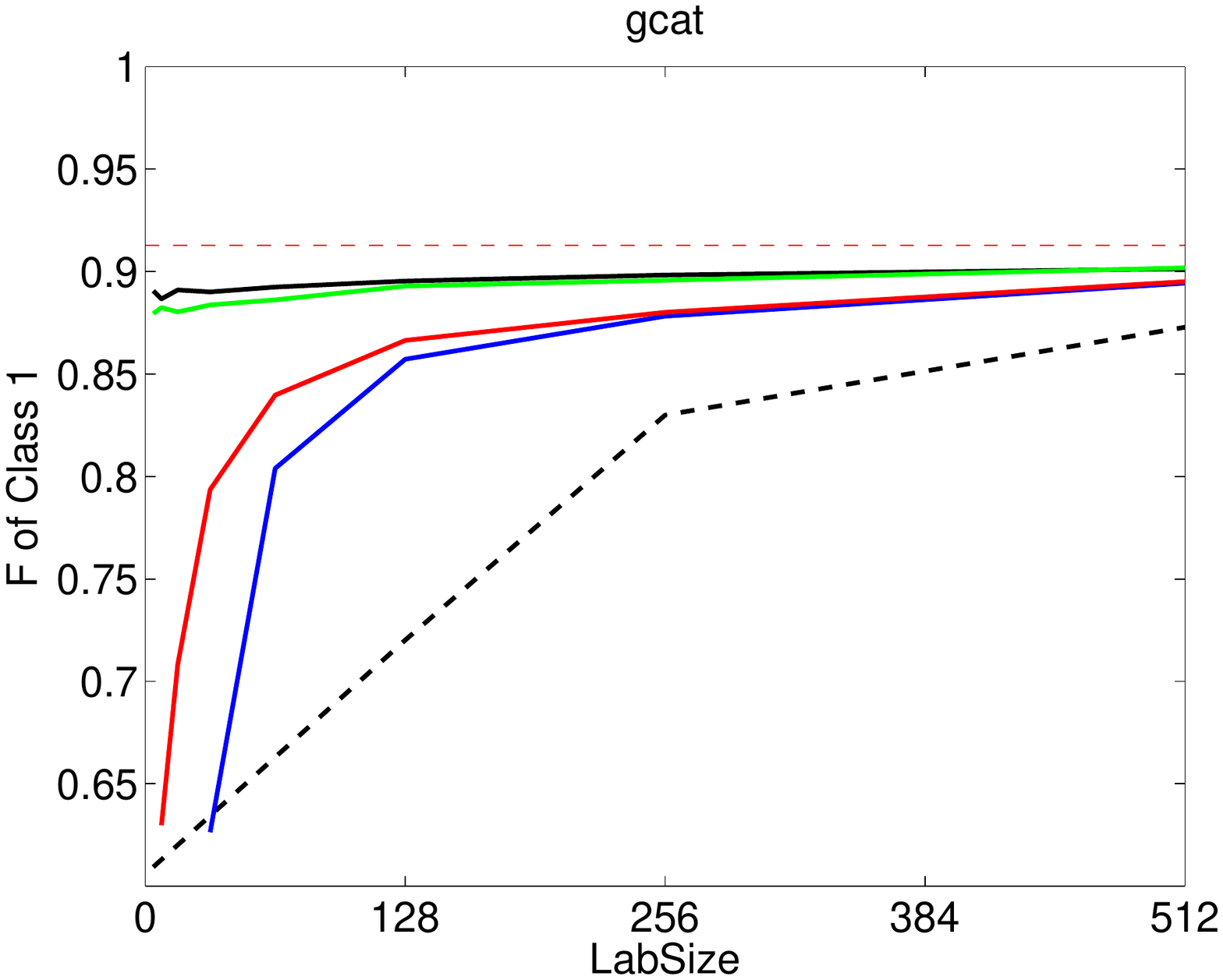}
   %\hspace{1cm}
   \includegraphics[width=0.45\linewidth]{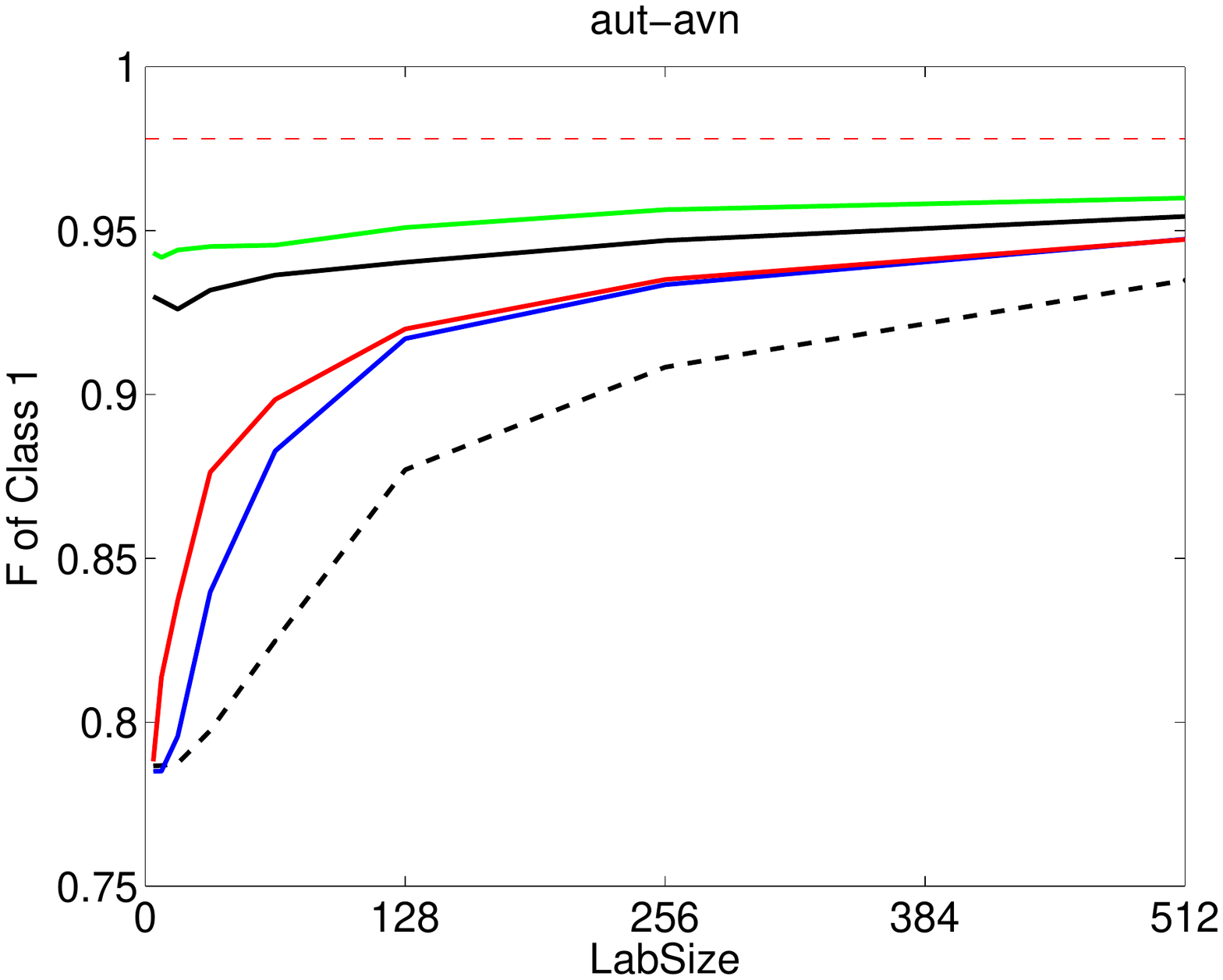} \\
\end{center}
\vspace*{-0.8in}
\caption{{Comparison of maxent methods on {\it gcat} and {\it aut-avn} datasets. Dashed Black: supervised learning,~\eqref{eq:sup}; Continuous Black: our method,~\eqref{eq:semisup1}-\eqref{eq:semisup2}; Green: entropy regularization,; Red: expectation constraint,; Blue: posterior regularization. Dashed horizontal red line corresponds to the supervised classifier built using $L$ and $U$ with their labels known.}}
\end{figure*}

\section{Maxent: Comparison with other semi-supervised methods}
\label{maxent}
%\vspace*{-0.1in}

One of the nice features of our method is its applicability to general loss functions. Here we take up the maxent loss, ~\eqref{eq:mentloss} and compare our method with other semi-supervised maxent methods which make use of domain constraints such as the label constraints in~\eqref{eq:semisup2}.
%\eat{
Let
\begin{eqnarray}
p_i^u(y_i^u) = \frac{\exp(\bw^T \bfe(y_i^u;\bx_i^u))}{\sum_{y} \exp(\bw^T \bfe(y;x_i^u))}, \nonumber \\
E_i^u = -\sum_{y_i^u} p_i^u(y_i^u) \log p_i^u(y_i^u) \nonumber
\end{eqnarray}
be, respectively, the probability of label $y_i^u$, the partition function, and the entropy of the label probability distribution associated with the $i$-th unlabeled example.
%}

%\begin{comment}
%\begin{figure}[h]
%%\label{fig:uv}
%\begin{center}
%%\framebox[4.0in]{$\;$}
%%\fbox{\rule[-.5cm]{0cm}{4cm} \rule[-.5cm]{4cm}{0cm}}
%\vspace*{-0.7in}
%   \includegraphics[width=0.4\linewidth]{vary_unlab_plots/Lcurve_F_20NG_uv.pdf}
%   \hspace{1cm}
%   \includegraphics[width=0.4\linewidth]{vary_unlab_plots/Lcurve_F_mnist_uv.pdf} \\
%\end{center}
%\vspace*{-0.9in}
%\caption{Multi-class datasets: Variation of performance (Macro F) as a function of the number of unlabeled examples (UnLabSize), with the number of labeled examples fixed at 80.}
%\end{figure}
%\end{comment}

\subsection{Entropy Regularization}
The method minimizes the following objective function:
\begin{eqnarray}
F^s(\bw) + C^u \sum_{i=1}^n E_i^u \;
{\rm s.t.} \; \sum_{i=1}^n p_i^u(y) = n(y) \dsp \forall y.
\label{eq:er}
\end{eqnarray}
%where $E_i^u=-\sum_{y} p_i^u(y) \log p_i^u(y)$, the entropy of the label probability distribution associated with the $i$-th unlabeled example.
Although the original entropy regularization method~\cite{Grandvalet2004} does not use the domain constraints in~\eqref{eq:er}, these constraints are crucial for getting good performance, and so we include them. The unlabeled data term in the objective function (which is referred to as the {\it entropy regularization} term), can be viewed as the expected negative log-likelihood of the label probability distribution on unlabeled data given by the model. This term can be compared with the unlabeled data term in the objective function associated with our formulation,~\eqref{eq:semisup1}. While we work with choosing a single label for each example, entropy regularization works with expectations. A key advantage of our method over entropy regularization is that the use of alternate optimization of $\bw$ and $\by^u$ on~\eqref{eq:semisup1}-\eqref{eq:semisup2} allows an easy handling of the domain constraints. This advantage can be particularly crucial when dealing with general structured prediction problems for which gradients of the domain constraint functions involving $p_i^u$ are expensive to compute~\cite{Jiao2006}.

\subsection{Expectation Regularization/Constraint}
\shortcite{Mann2010} use unlabeled data only to deal with the domain constraints; they solve the optimization problem,
$$\min_{\bw} \dsp F^s(\bw) + C^L \sum_{y=1}^m (\sum_{i=1}^n p_i^u(y)-n(y))^2.$$
If the $n(y)$ values are known precisely it is better to enforce the label constraints and solve, instead, the following problem:
%$$\min_{\bw} \; F^s(\bw)$ {\rm s.t.} $\sum_{i=1}^n p_i^u(y) = n(y) \; \forall y$$.
%\eat{
\begin{equation}
\min_{\bw} \dsp F^s(\bw) \dsp
{\rm s.t.} \dsp \sum_{i=1}^n p_i^u(y) = n(y) \dsp \forall y
\label{eq:lc}
\end{equation}
%}
Like entropy regularization, a disadvantage of this method is the need to deal with gradients of constraint functions involving $p_i^u$.

\subsection{Posterior Regularization}
This method~\cite{Gartner2005,Graca2007,Ganchev2009} was introduced mainly to ease the handling of constraints in the expectation regularization/constraint method.
This is achieved by introducing intermediate label distributions $q_i^y=\{q_i^u(y_i^u)\}_{y_i^u}\dsp\forall i$, forcing the constraints
%\eat{
\footnote{There is a minor difference with what is originally presented by~\cite{Gartner2005}, who include the labeled examples in the label constraints. But those equations can be rewritten in the form~\eqref{eq:pr} by appropriately defining $n(y)$.%}
 }
on $\{q_i^u\}$ and including a KL divergence term between $\{p_i^u\}$ and $\{q_i^u\}$:
%$\min_{\bw,\{q_i^u\}} F^s(\bw) + C^{KL} \sum_{i=1}^n KL(q_i^u,p_i^u)$
%{\rm s.t.} $\sum_{i=1}^n q_i^u(y) = n(y) \; \forall y$.
%\eat{
\begin{eqnarray}
\min_{\bw,\{q_i^u\}} F^s(\bw) + C^{KL} \sum_{i=1}^n
%\sum_{y_i^u} q_i^u(y_i^u) \log \frac{q_i^u(y_i^u)}{p_i^u(y_i^u)} \dsp
\sum_{y_i^u} q_i^u(y_i^u) \log \frac{q_i^u(y_i^u)}{p_i^u(y_i^u)} \nonumber \\
{\rm s.t.} \dsp \sum_{i=1}^n q_i^u(y) = n(y) \dsp \forall y \label{eq:pr}
\end{eqnarray}
%}
If alternating optimization is used on $\bw$ and $\{q_i^u\}$, then, like in our method, we only need to solve convex optimization problems in each step. We found $C^{KL}=0.1$ to be a good default value.

We implemented entropy regularization and expectation constraint methods, only for binary classification because of the complexity brought in by vector constraints. The augmented lagrangian method~\cite{Bertsekas1997} was used to handle the constraint. Posterior regularization was implemented as described in~\cite{Gartner2005}. Figure 4 compares the various methods on the two binary text classification datasets, {\it gcat} and {\it aut-avn}~\cite{Sindhwani2006}. {\it gcat} has 23149 examples and 47236 features; {\it aut-avn} has 71175 examples and 20707 features. The experimental set up is similar to that in section~\ref{expts} except: $L$ consists of 512 examples, and, performance was measured in terms of the F measure of the first class.

The performances of expectation constraint and posterior regularization methods are close, with the latter being slightly inferior due to the use of the intermediate distribution $q_i^u$ and alternate optimization. Both these methods are quite inferior to entropy regularization and our method; clearly, the unlabeled likelihood terms in~\eqref{eq:er} and~\eqref{eq:semisup1} play a crucial role in this. Our method is slightly inferior to entropy regularization due to the use of alternate optimization. All the four methods lift the performance of supervised learning quite well and so they are good semi-supervised techniques.

\section{Conclusion}
\label{concl}

%\vspace*{-0.1in}
In this paper we extended the TSVM approach of semi-supervised binary classification to multi-class and hierarchical classification problems with general loss functions, and demonstrated the effectiveness of the extended approach. As a natural next step we are exploring the approach for structured output prediction. The $\by^u$ determination process is harder in this case since reduction to linear programming is not automatic. But good solutions are still possible. In many applications of structured output prediction, labeled data consists of examples with partial labels. Our approach can easily handle this case; all that one has to do is include all unknown labels as a part of $\by^u$.

\newpage
\bibliographystyle{apa}

\bibliography{coling}

\begin{thebibliography}{}

\bibitem[\protect\astroncite{Bertsekas and Tsitsiklis}{1997}]{Bertsekas1997}
Bertsekas, D.~P. and Tsitsiklis, J.~N. (1997).
\newblock {\em Parallel and Distributed Computation: Numerical Methods}.
\newblock Athena Scientific.

\bibitem[\protect\astroncite{Bruzzone et~al.}{2006}]{Bruzzone2006}
Bruzzone, L., Chi, M., and Marconcini, M. (2006).
\newblock A novel transductive {SVM} for semisupervised classification of
  remote-sensing images.
\newblock volume~44, pages 3363--3373.

\bibitem[\protect\astroncite{Chang et~al.}{2007}]{Chang2007}
Chang, M.~W., Ratinov, L., and Roth, D. (2007).
\newblock Guiding semi-supervision with constraint-driven learning.
\newblock In {\em ACL}.

\bibitem[\protect\astroncite{Chapelle et~al.}{2006}]{Chapelle2006}
Chapelle, O., Chi, M., and Zien, A. (2006).
\newblock A continuation method for semi-supervised {SVM}s.
\newblock In {\em ICML}.

\bibitem[\protect\astroncite{Chapelle et~al.}{2008}]{Chapelle2008}
Chapelle, O., Sindhwani, V., and Keerthi, S.~S. (2008).
\newblock Optimization techniques for semi-supervised support vector machines.
\newblock In {\em JMLR}, volume~9, pages 203--233.

\bibitem[\protect\astroncite{De~Bie and Cristianini}{2004}]{Debie2004}
De~Bie, T. and Cristianini, N. (2004).
\newblock Convex methods for transduction.
\newblock In {\em NIPS}.

\bibitem[\protect\astroncite{Forman}{2003}]{Forman2003}
Forman, G. (2003).
\newblock An extensive empirical study of feature selection metrics for text
  classification.
\newblock In {\em JMLR}, volume~3, pages 1289--1305.

\bibitem[\protect\astroncite{Ganchev et~al.}{2009}]{Ganchev2009}
Ganchev, K., Graca, J., Gillenwater, J., and Taskar, B. (2009).
\newblock Posterior regularization for structured latent variable models.
\newblock Technical report, Dept. of Computer \& Information Science,
  University of Pennsylvania.

\bibitem[\protect\astroncite{G\"{a}rtner et~al.}{2005}]{Gartner2005}
G\"{a}rtner, T., Le, Q.~V., Burton, S., Smola, A.~J., and Vishwanathan, S.
  V.~N. (2005).
\newblock Large-scale multiclass transduction.
\newblock In {\em NIPS}.

\bibitem[\protect\astroncite{Graca et~al.}{2007}]{Graca2007}
Graca, J., Ganchev, K., and Taskar, B. (2007).
\newblock Expectation maximization and posterior constraints.
\newblock In {\em NIPS}.

\bibitem[\protect\astroncite{Grandvalet and Bengio}{2003}]{Grandvalet2004}
Grandvalet, Y. and Bengio, Y. (2003).
\newblock Semi-supervised learning by entropy minimization.
\newblock In {\em NIPS}.

\bibitem[\protect\astroncite{Hadley}{1963}]{Hadley1963}
Hadley, G. (1963).
\newblock {\em Linear Programming}.
\newblock Addison-Wesley, 2nd edition.

\bibitem[\protect\astroncite{Jiao et~al.}{2006}]{Jiao2006}
Jiao, J., Wang, S., Lee, S., Greiner, R., and Schuurmans, D. (2006).
\newblock Semi-supervised conditional random fields for improved sequence
  segmentation and labeling.
\newblock In {\em ACL}.

\bibitem[\protect\astroncite{Joachims}{1999}]{Joachims1999}
Joachims, T. (1999).
\newblock Transductive inference for text classification using support vector
  machines.
\newblock In {\em ICML}.

\bibitem[\protect\astroncite{Lang}{1995}]{Lang1995}
Lang, K. (1995).
\newblock Newsweeder: Learning to filter netnews.
\newblock In {\em ICML}.

\bibitem[\protect\astroncite{LeCun}{2011}]{LeCun}
LeCun, Y. (2011).
\newblock The {MNIST} database of handwritten digits.

\bibitem[\protect\astroncite{Lee et~al.}{2006}]{Lee2006}
Lee, C.~H., Wang, S., Jiao, F., Schuurmans, D., and Greiner, R. (2006).
\newblock Learning to model spatial dependency: semi-supervised discriminative
  random fields.
\newblock In {\em NIPS}.

\bibitem[\protect\astroncite{Lewis et~al.}{2006}]{Lewis2004}
Lewis, D., Yang, Y., Rose, T., and Li, F. (2006).
\newblock Rcv1: A new benchmark collection for text categorization research.
\newblock In {\em JMLR}, volume~5, pages 361--397.

\bibitem[\protect\astroncite{Mann and McCallum}{2010}]{Mann2010}
Mann, G.~S. and McCallum, A. (2010).
\newblock Generalized expectation criteria for semi-supervised learning with
  weakly labeled data.
\newblock In {\em JMLR}, volume~11, pages 955--984.

\bibitem[\protect\astroncite{McCallum and Nigam}{1998}]{McCallum1998}
McCallum, A. and Nigam, K. (1998).
\newblock A comparison of event models for naive {B}ayes text classification.
\newblock In {\em AAAI Workshop on Learning for Text Categorization}.

\bibitem[\protect\astroncite{Sindhwani and Keerthi}{2006}]{Sindhwani2006}
Sindhwani, V. and Keerthi, S. (2006).
\newblock Large-scale semi-supervised linear {SVM}s.
\newblock In {\em SIGIR}.

\bibitem[\protect\astroncite{Tibshirani}{2011}]{Tibshirani}
Tibshirani, R. (2011).
\newblock {USPS} handwritten digits dataset.

\bibitem[\protect\astroncite{Xu et~al.}{2006}]{Xu2006}
Xu, L., Wilkinson, D., Southey, F., and Schuurmans, D. (2006).
\newblock Discriminative unsupervised learning of structured predictors.
\newblock In {\em ICML}.

\bibitem[\protect\astroncite{Zien et~al.}{2007}]{Zien2007}
Zien, A., Brefeld, U., and Scheffer, T. (2007).
\newblock Transductive support vector machines for structured variables.
\newblock In {\em ICML}.

\bibitem[\protect\astroncite{Zubiaga et~al.}{2009}]{Zubiaga2009}
Zubiaga, A., Fresno, V., and Martinez, R. (2009).
\newblock Is unlabeled data suitable for multiclass {SVM}-based web page
  classification?
\newblock In {\em NAACL HLT Workshop on Semi-supervised Learning for Natural
  Language Processing}.

\end{thebibliography}
\end{document}